\documentclass[letterpaper, 10 pt, conference]{ieeeconf}  

\usepackage{graphicx}

\usepackage{dblfloatfix}

\usepackage{amsmath}

\usepackage{algorithm}

\usepackage{algpseudocode}

\usepackage{multirow}

\usepackage{booktabs}

\usepackage{tabularx}

\usepackage{array}

\usepackage{makecell}

\usepackage{amssymb}

\usepackage{bm}

\usepackage{upgreek}

\newcolumntype{Y}{>{\centering\arraybackslash}X}

\newcolumntype{C}[1]{%
  >{\centering\arraybackslash}m{#1}%
}

\IEEEoverridecommandlockouts                              

\title{\LARGE \bf
M2-SMap: Memory-Efficient Semantic Mapping with

Hierarchical Multi-Model Representation
}

\author{QiYing Deng$^{1}$, ZhongLai Wang$^{1,*}$, Yuan Gao$^{2}$, Wei Dong$^{2}$
\thanks{*This work was supported by the AAAA. (Corresponding author: ZhongLai Wang.)}
\thanks{$^{1}$QiYing Deng and ZhongLai Wang are with University of Electronic Science and Technology of China (e-mail: 202422040355@std.uestc.edu.cn; wzhonglai@uestc.edu.cn).}
\thanks{$^{2}$Yuan Gao and Wei Dong are with the State Key Laboratory of Mechanical System and Vibration, School of Mechanical Engineering, Shanghai Jiao Tong University, China (e-mail: GaoY-23@sjtu.edu.cn; dr.dongwei@sjtu.edu.cn).}
}

\begin{document}

\maketitle
\thispagestyle{empty}
\pagestyle{empty}


\begin{abstract}
Dense point cloud maps, as a typically used mapping representation, are difficult to deploy on resource-constrained robots because their memory consumption grows rapidly with scene scale.
Although compact single-model representations reduce memory cost, their fixed geometric expressiveness is insufficient for structurally diverse environments. Existing multi-model methods improve representational flexibility, yet their feature extraction and model selection are often dominated by local geometry, which can cause overfitting and adhesion between objects.
To address these issues, this paper presents M2-SMap, a memory-efficient semantic mapping framework based on hierarchical multi-model representation.
First, a hierarchical geometric decomposition partitions RGB-D point clouds into compact Gaussian components.
Then, a projection-guided semantic annotation mechanism assigns instance identities to each component.
Subsequently, these annotations are incorporated into an object-aware Gaussian fusion strategy.
Furthermore, a multi-scale feature extraction strategy separates large planar regions, semantic objects, and complex residual structures, which are respectively represented by bounded planes, object-level superquadrics, and GMM primitives.
Experiments on three RGB-D sequences show that M2-SMap runs in real time at no less than 29.37~Hz while achieving the lowest primitive count, with an average reduction of 18.7\% over the best baseline. It also reduces the mean per-frame number of measured inter-object adhesion cases from 2.808 to 0, demonstrating efficient and semantically consistent scene representation.

\end{abstract}

\section{INTRODUCTION}

3D scene mapping provides the geometric foundation for robotic localization, navigation, and interaction~\cite{hornung2013octomap,oleynikova2017voxblox,rusu2008objectmaps,mccormac2017semanticfusion}. Point clouds directly preserve the structure observed by RGB-D cameras, but their storage requirements increase substantially as the mapped environment expands. Therefore, resource-constrained robots require map representations that remain both compact and computationally efficient without sacrificing essential geometric expressiveness.

\begin{figure}[t]
\centering
\includegraphics[width=0.98\columnwidth]{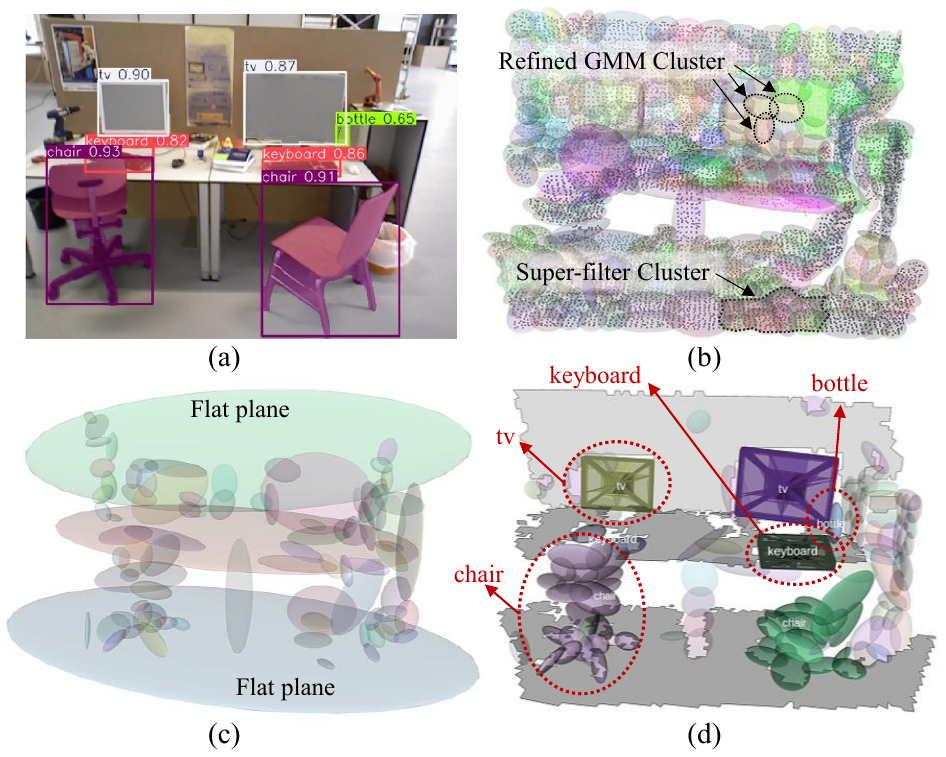}
\caption{Overall results of the proposed M2-SMap framework. (a) Original RGB image with instance segmentation. (b) Overlay of supervoxelized point-cloud clusters and the refined GMMs. (c) Result after GMM fusion. (d) Final multi-model semantic representation.}
\label{fig:motivation}
\end{figure}

Traditional parametric mapping methods are mainly based on single-model representations, which often struggle to satisfy these requirements simultaneously. Voxel, octree, TSDF, and ESDF methods support online mapping through spatial discretization or signed-distance estimation~\cite{hornung2013octomap,newcombe2011kinectfusion,oleynikova2017voxblox}, while their accuracy remains closely coupled with map resolution, creating an accuracy--memory trade-off. Plane-based methods compactly model large planar regions~\cite{poppinga2008fastplane}, whereas surfel maps provide flexible local surface reconstruction~\cite{whelan2015elasticfusion} but retain many local elements in large or geometrically complex scenes. B-spline surfaces represent complex geometry with few parameters, although merging overlapping patches is nontrivial~\cite{yan2015bspline}. Quadrics compactly encode object position, orientation, and extent~\cite{nicholson2018quadricslam,liao2020rgbdquadric}, but their ellipsoidal form restricts the shapes represented by a single primitive. Gaussian Mixture Models (GMMs) replace dense local points with probabilistic components parameterized by means and covariance matrices~\cite{srivastava2019hierarchicalgmm,dhawale2020surfacegmm,li2022spgf}, while complex boundaries may still require multiple low-degree-of-freedom components~\cite{gao2025multimodel}. Deep implicit representations reduce explicit storage at the expense of costly optimization and GPU acceleration, which limits their use on low-power platforms~\cite{sucar2021imap,zhu2022niceslam}. Consequently, the fixed geometric expressiveness of a single model makes it difficult to represent diverse scenes with bounded memory and computation.

\begin{figure*}[!t]
    \centering
    \includegraphics[width=\textwidth]{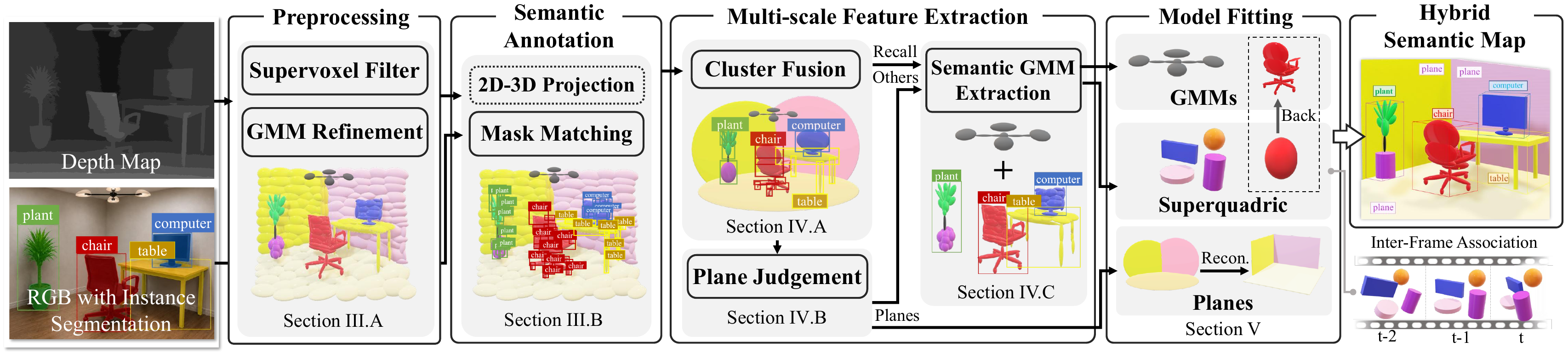}
    \caption{Overview of M2-SMap. RGB-D observations are converted into refined Gaussian components and annotated through 3D-to-2D projection and mask matching. Multi-scale feature extraction then produces the final hierarchical representation using planes, superquadrics, and GMMs.}
    \label{fig:overview}
\end{figure*}

To overcome the limitations of single-model representations, multi-model methods combine models with complementary geometric properties and assign them to different scene regions. Gao and Dong integrate planes, GMMs, and B-spline surfaces for real-time point-cloud representation~\cite{gao2025multimodel}, while PLC-LiSLAM jointly optimizes planes, lines, and cylinders in LiDAR SLAM~\cite{zhou2022plclislam}. These methods improve geometric flexibility, but compact maps for resource-constrained systems must still limit the number of model instances. Therefore, each model may cover a relatively large or geometrically mixed region, and model selection based mainly on local connectivity or fitting residuals can merge adjacent objects or fragment one object with several models. This causes blind model allocation and inter-object adhesion in complex scenes.

The limitations above indicate that compact multi-model mapping requires semantics not only to describe the scene but also to guide geometric allocation. Existing semantic mapping methods can be broadly grouped according to how semantic information interacts with geometry. The first group annotates a predefined 3D representation. Early RGB-D systems fuse class predictions into reconstructed maps~\cite{stueckler2012semantic,stueckler2015dense,hermans2014dense}. SemanticFusion integrates CNN predictions into an ElasticFusion surfel map~\cite{mccormac2017semanticfusion}, while PanopticFusion maintains semantic classes and instance identities in a spatially hashed TSDF volume~\cite{narita2019panopticfusion}. These methods improve persistent labeling and multi-view semantic consistency. However, the annotated primitives are still determined by geometry alone, so semantic labels neither correct geometry-based grouping nor guide compact parametric model selection.

The second group uses semantic instances to allocate separate geometric models. MaskFusion maintains independent surfel models for segmented objects~\cite{runz2018maskfusion}, Voxblox++ constructs volumetric object-centric maps~\cite{grinvald2019voxbloxpp}, Fusion++ initializes per-object TSDF reconstructions~\cite{mccormac2018fusionpp}, and Panoptic Multi-TSDFs assigns separate submaps and class-dependent resolutions~\cite{schmid2022panopticmultitsdf}. Such object-wise allocation can reduce cross-object fusion, but it relies on dense surfel or TSDF submaps together with instance or panoptic segmentation. MaskFusion was evaluated using two Titan X GPUs~\cite{runz2018maskfusion}, Voxblox++ operated at 1~Hz in its online robotic experiment~\cite{grinvald2019voxbloxpp}, and Fusion++ reported 4--8~Hz~\cite{mccormac2018fusionpp}. The cost of dense model maintenance and learned segmentation makes these systems difficult to deploy on resource-constrained RGB-D platforms and leaves a need for lightweight semantic guidance of compact primitive allocation.

To address blind model selection and inter-object adhesion, this paper proposes M2-SMap, a semantic mapping framework based on hierarchical adaptive multi-model fusion for resource-constrained RGB-D robots, as illustrated in Fig.~\ref{fig:overview}. M2-SMap validates instance annotations on refined Gaussian components before Gaussian fusion and flat-plane separation. The support points of each component are projected into the registered image and matched with instance masks, after which the validated instance attributes constrain geometric merging. Large flat regions, regular semantic objects, and complex residual structures are then represented by bounded planes, object-level superquadrics, and GMMs, respectively. This hierarchy combines compact object-level abstraction with local geometric flexibility. The complete pipeline is shown in Fig.~\ref{fig:overview}.

The main contributions are as follows.

(1) A semantic-guided hierarchical mapping framework combines bounded planes, object-level superquadrics, and GMMs for compact RGB-D scene representation.

(2) Instance-mask annotations are introduced before Gaussian fusion and flat-plane separation, reducing irreversible inter-object adhesion while preserving object boundaries.


(3) Our method significantly improves representation compactness. Compared with the best existing baseline, it reduces the average primitive count by 18.7\%.

\section{ALGORITHM OVERVIEW}
The overall pipeline of M2-SMap is shown in Fig.~\ref{fig:overview} and summarized in Algorithm~\ref{alg:m2smap}. Starting from each RGB-D frame, the algorithm performs hierarchical Gaussian decomposition, projection-guided instance annotation, and semantic-constrained Gaussian fusion. It then separates flat planes, semantic objects, and residual structures, which are represented by planes, superquadrics, or GMM primitives.

\begin{algorithm}[t]
\caption{Semantic-Constrained Hierarchical Mapping}
\label{alg:m2smap}
\footnotesize
\algrenewcommand\algorithmicindent{1em}
\begin{algorithmic}[1]

\Require Registered RGB image $\mathbf{I}_t$, depth image
$\mathbf{D}_t$, instance-mask set $\mathcal{M}_t$, camera intrinsic
matrix $\mathbf{K}$, previous superquadric tracks
$\mathcal{T}^{sq}_{t-1}$, and mapping parameters
$\bm{\Theta}$

\Ensure Hierarchical primitive map $\mathcal{H}_t$ and updated
superquadric tracks $\mathcal{T}^{sq}_t$

\State $\mathcal{P} \gets
\Call{GeneratePointCloud}
{\mathbf{I}_t,\mathbf{D}_t,\mathbf{K}}$

\State $\mathcal{G}_{\mathrm{raw}} \gets
\Call{HierarchicalSegmentation}{\mathcal{P}}$

\State $\mathcal{G}_{\mathrm{sem}} \gets \varnothing$

\ForAll{$G_i\in\mathcal{G}_{\mathrm{raw}}$}

    \State $\mathcal{U}_i \gets
    \Call{ProjectToImage}
    {G_i,\mathbf{K}}$

    \State $\mathbf{M}_i^{*} \gets
    \Call{FindBestMaskMatch}
    {\mathcal{U}_i,\mathcal{M}_t}$

    \State $G_i^{s} \gets
    \Call{AssignInstanceSemantics}
    {G_i,\mathbf{M}_i^{*}}$

    \State $\mathcal{G}_{\mathrm{sem}}
    \gets
    \mathcal{G}_{\mathrm{sem}}\cup\{G_i^{s}\}$

\EndFor

\State $\mathcal{G}_{\mathrm{fused}} \gets \varnothing$

\ForAll{candidate pairs
$(G_i,G_j)\subseteq\mathcal{G}_{\mathrm{sem}}$}

    \algrenewcommand\algorithmicthen{}

    \If{%
        \begin{tabular}[t]{@{}l@{}}
            $\Call{GeometricallyConsistent}{G_i,G_j}$\\[-0.15em]
            \makebox[0pt][r]{\textbf{and}\enspace}%
            $\Call{SemanticallyConsistent}{G_i,G_j}$ \textbf{then}
        \end{tabular}%
    }

    \algrenewcommand\algorithmicthen{\textbf{then}}

        \State $\mathcal{G}_{\mathrm{fused}}
        \gets
        \mathcal{G}_{\mathrm{fused}}
        \cup
        \Call{Fuse}{G_i,G_j}$

    \Else
        \State $\Call{RetainOriginalComponents}{G_i,G_j}$
    \EndIf

\EndFor

\State $\mathcal{P}_{\mathrm{plane}}\gets\varnothing$

\ForAll{$G_i^{f}\in\mathcal{G}_{\mathrm{fused}}$}

    \If{$\Call{SatisfyPlaneConditions}{G_i^{f}}$}

        \State $\mathcal{P}_{\mathrm{plane}}
        \gets
        \mathcal{P}_{\mathrm{plane}}
        \cup
        \Call{ReconstructPlane}{G_i^{f}}$

    \EndIf

\EndFor

\State $\mathcal{G}_{\mathrm{nonplane}} \gets
\Call{RemovePlaneRelatedGaussians}
{\mathcal{G}_{\mathrm{raw}},
 \mathcal{P}_{\mathrm{plane}}}$

\State $(\mathcal{O}_{\mathrm{sem}},
\mathcal{G}_{\mathrm{unknown}})
\gets
\Call{ConstructObjectInstances}
{\mathcal{G}_{\mathrm{nonplane}}}$

\ForAll{$\mathcal{O}_k\in\mathcal{O}_{\mathrm{sem}}$}

    \State $Q_k \gets
    \Call{FitSuperquadric}{\mathcal{O}_k}$

    \If{$\Call{IsValidSuperquadric}{Q_k}$}

        \State $(h_k,Q_k^{a}) \gets
        \Call{AssociateAndAlignSQ}
        {Q_k,\mathcal{T}^{sq}_{t-1},\bm{\Theta}}$

        \State $Q_k \gets
        \Call{SmoothSuperquadric}{Q_k^{a},h_k,\bm{\Theta}}$

        \State $\Call{UpdateSuperquadricMap}{Q_k}$

    \Else

        \State $\mathcal{G}_k \gets
        \Call{HierarchicalGaussianRefinement}{\mathcal{O}_k}$

        \State $\Call{UpdateGaussianMap}{\mathcal{G}_k}$

    \EndIf

\EndFor

\State $\Call{UpdateGaussianMap}
{\mathcal{G}_{\mathrm{unknown}}}$

\State $\Call{UpdatePlaneMap}
{\mathcal{P}_{\mathrm{plane}}}$

\State \Return updated hierarchical primitive map

\end{algorithmic}
\end{algorithm}

\section{Geometric Refinement and Mask-Guided Gaussian Annotation}

\subsection{Hierarchical Geometric Refinement}

A registered RGB-D frame is first converted into a point cloud, which is then sampled and pre-segmented into supervoxels. These supervoxels form locally coherent geometric units that preserve surface continuity while reducing the influence of depth noise.

\begin{figure}[t]
\centering
\includegraphics[width=0.98\columnwidth]{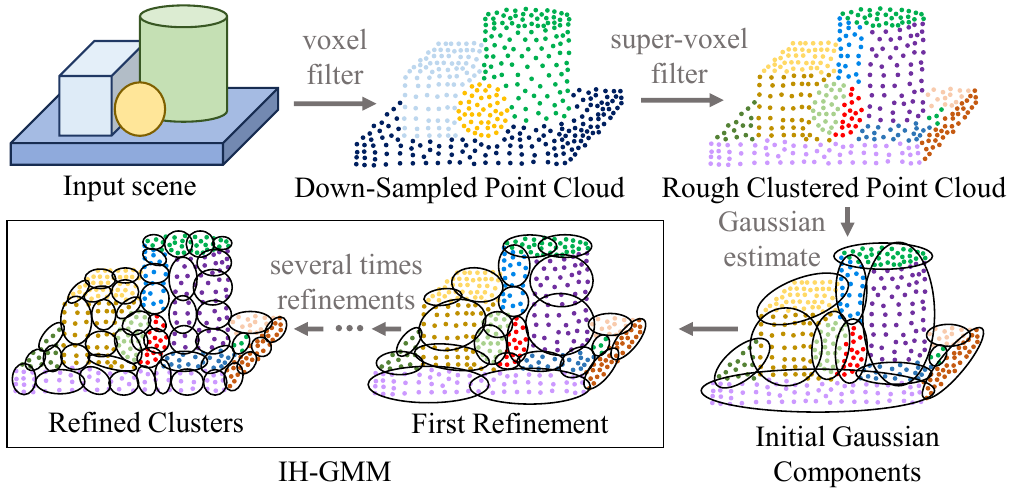}
\caption{Hierarchical geometric refinement. The filtered point cloud is partitioned into locally coherent supervoxels, whose point clusters are further decomposed into compact Gaussian components using the adopted IH-GMM refinement.}
\label{fig:hierarchical_refinement}
\end{figure}

Each supervoxel produced by the preceding pre-segmentation defines an input point cluster $\mathcal{C}_i=\{\mathbf{p}_{ik}\}_{k=1}^{N_i}$ before hierarchical Gaussian refinement. Its mean and covariance are $\bm{\upmu}_i=\frac{1}{N_i}\sum_{k=1}^{N_i}\mathbf{p}_{ik}$ and $\bm{\Sigma}_i=\frac{1}{N_i-1}\sum_{k=1}^{N_i}(\mathbf{p}_{ik}-\bm{\upmu}_i)(\mathbf{p}_{ik}-\bm{\upmu}_i)^{T}$, respectively. M2-SMap adopts the integrated hierarchical GMM (IH-GMM) refinement introduced by Gao and Dong~\cite{gao2023integratedgmm} and subsequently used in their multi-model framework~\cite{gao2025multimodel} to decompose each input cluster into compact Gaussian components, as illustrated in Fig.~\ref{fig:hierarchical_refinement}. The resulting components form the original Gaussian set $\mathcal{C}_{g}$.

\begin{figure}[t]
\centering
\includegraphics[width=0.98\columnwidth]{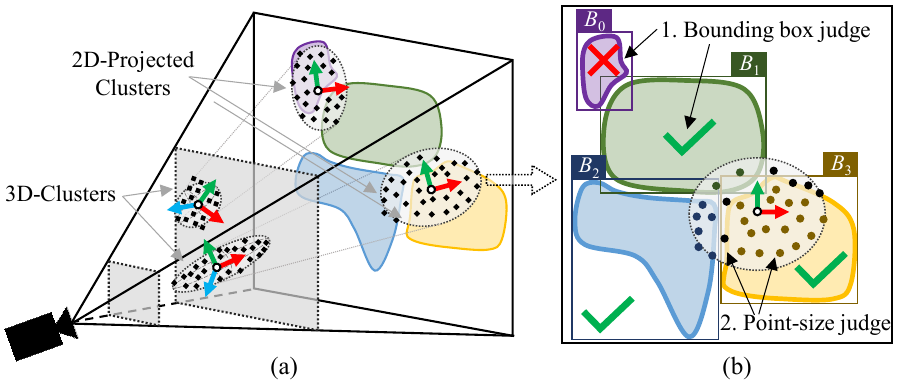}
\caption{Projection-guided instance-mask matching. (a) The 3D support points of each refined Gaussian component are projected into the registered image. (b) Non-overlapping instance candidates are rejected using their bounding boxes, and the remaining masks are ranked by the valid projected-point support ratio.}
\label{fig:projection_mask_matching}
\end{figure}

\subsection{Gaussian-to-Image Projection and Mask Matching}

Semantic annotations are assigned to the refined Gaussian components by projecting their supporting 3D points onto the corresponding RGB image. For each Gaussian component, its valid projected pixels are evaluated against the instance masks predicted by a lightweight instance-segmentation model. As shown in Fig.~\ref{fig:projection_mask_matching}, the instance bounding boxes first exclude clearly non-overlapping candidates, after which mask-level matching is performed according to the proportion of projected support points falling inside each candidate mask. The matched instance ID, category label, and detection confidence are stored as semantic attributes of the Gaussian component.

Let the instance-segmentation output be $\mathcal{Y}=\{(\mathbf{M}_j,\mathcal{B}_j^{\mathrm{2D}},\ell_j,s_j)\}_{j=1}^{N_y}$, where $N_y$ is the number of detected instances, $\mathbf{M}_j$ is the binary mask, $\mathcal{B}_j^{\mathrm{2D}}$ is its 2D bounding box, $\ell_j$ is the category label, and $s_j$ is the detection confidence. For a point $\mathbf{p}_{ik}^{c}=(x_{ik},y_{ik},z_{ik})^{T}$ in Gaussian component $\mathcal{C}_i$, its image coordinate is obtained inline as $\mathbf{q}_{ik}=\pi_{\mathbf{K}}(\mathbf{p}_{ik}^{c})=[u_{ik},v_{ik}]^{T}=[f_x x_{ik}/z_{ik}+c_x,\ f_y y_{ik}/z_{ik}+c_y]^{T}$. Projected points inside the image define the valid index set $\mathcal{V}_i=\{k\mid 0\leq\operatorname{round}(u_{ik})<W,\ 0\leq\operatorname{round}(v_{ik})<H\}$, where $W$ and $H$ are the image width and height. The quantity $N_i^{\mathrm{proj}}=|\mathcal{V}_i|$ is the number of valid projected points from component $\mathcal{C}_i$.

Let $u_i^{\min}$, $u_i^{\max}$, $v_i^{\min}$, and $v_i^{\max}$ be the coordinate extrema over $k\in\mathcal{V}_i$. The projected Gaussian support is the 2D box set $\mathcal{B}_i^{\mathrm{2D}}=[u_i^{\min},u_i^{\max}]\times[v_i^{\min},v_i^{\max}]$. It is used only to reject non-overlapping masks through $\mathcal{J}_i=\{j\mid s_j\geq\tau_s,\ \mathcal{B}_i^{\mathrm{2D}}\cap\mathcal{B}_j^{\mathrm{2D}}\neq\varnothing\}$. Pixel-level mask matching is then performed only for candidates in $\mathcal{J}_i$.

The remaining candidates are ranked by confidence-weighted projected-point support. When the best match satisfies the minimum projection, confidence, and support-ratio requirements, the component stores its instance ID $a_i$, label $\ell_i$, confidence $s_i$, and mask-support ratio $\rho_i$. Otherwise, $a_i=-1$ and the component remains semantically unassigned.

\begin{figure}[t]
\centering
\includegraphics[width=0.98\columnwidth]{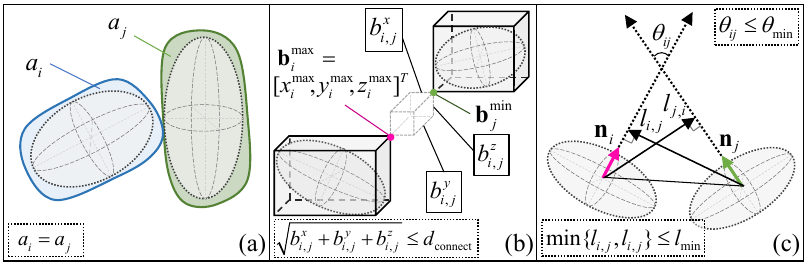}
\caption{Semantic-constrained Gaussian fusion. A candidate pair is merged only when it satisfies (a) instance-ID consistency $a_i=a_j$, (b) AABB connectivity $d_{i,j}^{\mathrm{box}}\leq d_{\mathrm{connect}}$, and (c) normal-angle and bidirectional point-to-plane constraints $\theta_{i,j}\leq\theta_{\min}$ and $\min(l_{i,j},l_{j,i})\leq l_{\min}$.}
\label{fig:semantic_gaussian_fusion}
\end{figure}

\section{Multi-scale Feature Extraction}

M2-SMap extracts complementary geometric features at two coupled scales. Fused clusters reveal large planar structures for flat-plane judgment, while the preserved fusion mapping returns to the original Gaussian scale to retain semantic objects and residual geometric details.

\subsection{Semantic-Constrained Gaussian Fusion}

Let $\mathcal{C}_i$ and $\mathcal{C}_j$ be two annotated Gaussian components with instance IDs $a_i$ and $a_j$. The semantic condition in Fig.~\ref{fig:semantic_gaussian_fusion}(a) is $\psi_s(\mathcal{C}_i,\mathcal{C}_j)=\mathbb{I}(a_i=a_j)$, where $\mathbb{I}(\cdot)$ is the indicator function, equal to $1$ when its argument is true and $0$ otherwise. Thus, $a_i=a_j=-1$ permits fusion between two unassigned components, whereas an object--plane pair or two different object instances are rejected.

For geometric connectivity, define the 3D axis-aligned bounding box of $\mathcal{C}_i$ as the set $\mathcal{B}_i^{\mathrm{3D}}=\prod_{r\in\{x,y,z\}}[b_{i,r}^{\min},b_{i,r}^{\max}]$. For $r\in\{x,y,z\}$, the non-overlapping interval gap between $\mathcal{B}_i^{\mathrm{3D}}$ and $\mathcal{B}_j^{\mathrm{3D}}$ is $b_{i,j}^{r}=\max(0,\,b_{j,r}^{\min}-b_{i,r}^{\max},\,b_{i,r}^{\min}-b_{j,r}^{\max})$, and the 3D box distance shown in Fig.~\ref{fig:semantic_gaussian_fusion}(b) is
\begin{equation}
d_{i,j}^{\mathrm{box}}=
\sqrt{(b_{i,j}^{x})^2+(b_{i,j}^{y})^2+(b_{i,j}^{z})^2}
\leq d_{\mathrm{connect}}.
\label{equ_bbox_connectivity}
\end{equation}
Here, $d_{\mathrm{connect}}$ is the maximum admissible separation between two component boxes.

Let $\mathbf{n}_i$ and $\mathbf{n}_j$ be the least-variance eigenvectors of the two covariance matrices. Their unsigned normal angle is $\theta_{i,j}=\arccos(|\mathbf{n}_i^{T}\mathbf{n}_j|)$, and fusion requires $\theta_{i,j}\leq\theta_{\min}$, where $\theta_{\min}$ denotes the maximum allowed angular discrepancy in the notation of Fig.~\ref{fig:semantic_gaussian_fusion}(c). The bidirectional centroid-to-plane distances are $l_{i,j}=|(\bm{\upmu}_j-\bm{\upmu}_i)^{T}\mathbf{n}_i|$ and $l_{j,i}=|(\bm{\upmu}_i-\bm{\upmu}_j)^{T}\mathbf{n}_j|$. The coplanarity test is $\min(l_{i,j},l_{j,i})\leq l_{\min}$. Consequently, the geometric predicate satisfies $\psi_g(\mathcal{C}_i,\mathcal{C}_j)=1$ only when $d_{i,j}^{\mathrm{box}}\leq d_{\mathrm{connect}}$, $\theta_{i,j}\leq\theta_{\min}$, and $\min(l_{i,j},l_{j,i})\leq l_{\min}$. Otherwise, $\psi_g(\mathcal{C}_i,\mathcal{C}_j)=0$. The complete fusion condition is
\begin{equation}
\psi(\mathcal{C}_i,\mathcal{C}_j)=\psi_g(\mathcal{C}_i,\mathcal{C}_j)\wedge\psi_s(\mathcal{C}_i,\mathcal{C}_j).
\label{equ_semantic_fusion}
\end{equation}
A connectivity graph is constructed from $\psi(\mathcal{C}_i,\mathcal{C}_j)$. Its connected components are merged into larger clusters while preserving the mapping $\Gamma$ from each fused cluster to its original Gaussian components.

\subsection{Flat-Plane Separation}

\begin{figure}[t]
\centering
\includegraphics[width=0.98\columnwidth]{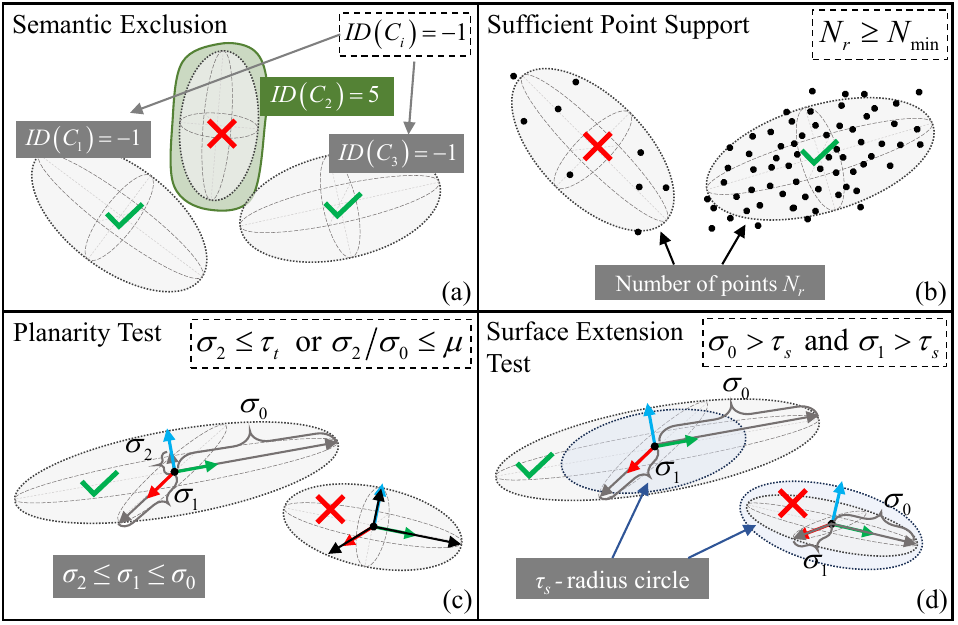}
\caption{Semantic-constrained flat-plane judgment. A fused cluster is accepted only when it has no valid object-instance ID, sufficient point support, a small thickness or principal-scale ratio, and sufficient extension along both principal directions.}
\label{fig:flat_plane_judgment}
\end{figure}

After semantic-constrained fusion, M2-SMap identifies large flat-plane candidates from the fused clusters. Let the covariance eigenvalues of a fused cluster $\mathcal{C}_i$ be $\lambda_{i,0}\geq\lambda_{i,1}\geq\lambda_{i,2}$, with principal scales $\sigma_{i,k}=\sqrt{\lambda_{i,k}}$. The conditions illustrated in Fig.~\ref{fig:flat_plane_judgment} are defined as follows. First, $N_i=|\mathcal{C}_i|\geq N_{\min}$ provides sufficient point support. Second, the cluster is planar when $\sigma_{i,2}\leq\tau_t$ or $\sigma_{i,2}/\sigma_{i,0}\leq\mu$, where $\tau_t$ is the thickness threshold. The implementation evaluates the equivalent eigenvalue-ratio form $\lambda_{i,2}/\lambda_{i,0}\leq\tau_{\lambda}$, hence $\mu=\sqrt{\tau_{\lambda}}$. Third, sufficient extension along both principal directions requires $\sigma_{i,0}\geq\tau_s$ and $\sigma_{i,1}\geq\tau_s$, where $\tau_s$ is the minimum principal scale.

Let $\mathrm{ID}(\mathcal{C}_i)=a_i$ denote the validated instance ID carried by the fused cluster. The complete flat-plane indicator is
\begin{equation}
\begin{aligned}
\chi_p(\mathcal{C}_i)=\mathbb{I}\bigg(&N_i\geq N_{\min}\ \wedge\\
&\left(\sigma_{i,2}\leq\tau_t\ \vee\ 
\frac{\sigma_{i,2}}{\sigma_{i,0}}\leq\mu\right)\ \wedge\\
&\sigma_{i,0}\geq\tau_s\ \wedge\ \sigma_{i,1}\geq\tau_s\ \wedge\\
&\mathrm{ID}(\mathcal{C}_i)=-1\bigg).
\end{aligned}
\label{equ_semantic_plane_condition}
\end{equation}
A locally planar cluster associated with an object instance therefore remains in the object branch, preventing thin object surfaces such as monitors and laptops from being removed as flat-plane support.

\subsection{Semantic GMM Extraction}

The fused clusters are used only for large-scale planar judgment. Object and residual modeling returns to the original Gaussian scale. For each accepted plane support $\mathcal{P}_r$, the fusion mapping $\Gamma$ traces its source fused cluster back to the corresponding original Gaussian indices $\mathcal{G}(\mathcal{P}_r)$. Let $\mathcal{G}_{o}$ be the original Gaussian components carrying validated object-instance annotations. The confirmed plane-related indices are $\mathcal{G}_{\mathrm{pl}}=\bigcup_{\mathcal{P}_r\in\mathcal{P}_{\mathrm{acc}}}(\mathcal{G}(\mathcal{P}_r)\setminus\mathcal{G}_{o})$, and the retained non-plane Gaussian set is $\mathcal{C}_{np}=\{\mathcal{C}_k^g\in\mathcal{C}_g\mid k\notin\mathcal{G}_{\mathrm{pl}}\}$. This mapping-based extraction removes original GMMs confirmed as flat-plane support while preserving semantic GMMs and unmatched residual GMMs.

The retained components are divided into the semantic set $\mathcal{C}_{sem}=\{\mathcal{C}_i^g\in\mathcal{C}_{np}\mid a_i\geq0\}$ and the residual set $\mathcal{C}_{res}=\{\mathcal{C}_i^g\in\mathcal{C}_{np}\mid a_i=-1\}$. Components in $\mathcal{C}_{sem}$ with the same valid instance ID are aggregated into the object-level support set $\mathcal{O}_j=\bigcup_{i:a_i=j,\,\mathcal{C}_i^g\in\mathcal{C}_{sem}}\mathcal{C}_i^g$. Each object instance stores its category label, confidence, mask-support statistics, associated bounding box, original Gaussian indices, and aggregated 3D support. Components in $\mathcal{C}_{res}$ remain local GMM candidates for irregular or unmatched structures in Section~V.

\section{Hierarchical Adaptive Model Fitting}

Flat-plane supports are reconstructed as bounded planes. Semantic object supports are first fitted with superquadrics and fall back to their original GMM components when a valid fit cannot be obtained. Residual components are represented by GMM primitives.

\subsection{Plane Reconstruction}

For an accepted plane support $\mathcal{P}_r$, the centroid $\bm{\upmu}_r$ and the first two covariance eigenvectors $\mathbf{e}_{r,1}$ and $\mathbf{e}_{r,2}$ define the local tangent frame. The unit normal is $\mathbf{n}_r=\mathbf{e}_{r,1}\times\mathbf{e}_{r,2}$, and the analytic support plane satisfies $\mathbf{n}_r^{T}(\mathbf{x}-\bm{\upmu}_r)=0$. Each support point $\mathbf{p}\in\mathcal{P}_r$ is mapped to local coordinates $u(\mathbf{p})=(\mathbf{p}-\bm{\upmu}_r)^{T}\mathbf{e}_{r,1}$ and $v(\mathbf{p})=(\mathbf{p}-\bm{\upmu}_r)^{T}\mathbf{e}_{r,2}$. For a grid size $g_p$, the occupied-cell set is
\begin{equation}
\mathcal{H}_r=\left\{
\left(\left\lfloor\frac{u(\mathbf{p})}{g_p}\right\rfloor,
\left\lfloor\frac{v(\mathbf{p})}{g_p}\right\rfloor\right)
\;\middle|\;\mathbf{p}\in\mathcal{P}_r
\right\}.
\label{equ_plane_occupied_cells}
\end{equation}
One representative point is reconstructed at the center of each occupied cell $(a,b)\in\mathcal{H}_r$:
\begin{equation}
\hat{\mathbf{p}}_{r,ab}=\bm{\upmu}_r+
\left(a+\frac{1}{2}\right)g_p\mathbf{e}_{r,1}+
\left(b+\frac{1}{2}\right)g_p\mathbf{e}_{r,2}.
\label{equ_plane_reconstruction}
\end{equation}
The plane primitive stores this occupied support together with its local frame, finite extent, thickness, eigenvalue ratio, color, and source cluster ID. Reconstructing only occupied cells keeps the plane bounded by the observed support and avoids filling unobserved regions inside a complete rectangle.

\subsection{Superquadric Fitting for Semantic Objects}

For a semantic object support $\mathcal{O}_i$, M2-SMap uses a superquadric as a compact analytic model. Its scale, shape, translation, and rotation parameters are collected in the vector $\bm{\Theta}^{\mathrm{sq}}=[s_x,s_y,s_z,\epsilon_1,\epsilon_2,t_x,t_y,t_z,r_x,r_y,r_z]^{T}$. Given a point $\mathbf{p}_k$, its local coordinate is $\tilde{\mathbf{p}}_k=\mathbf{R}^{T}(\mathbf{p}_k-\mathbf{t})$. The implementation uses the residual
\begin{equation}
r_k(\bm{\Theta}^{\mathrm{sq}})=\sqrt{H(\tilde{\mathbf{p}}_k,\bm{\Theta}^{\mathrm{sq}})}-1,
\label{equ_sq_residual}
\end{equation}
where $H(\cdot)$ is the superquadric implicit function. The fitting objective is
\begin{equation}
E_{sq}=\frac{1}{|\mathcal{O}_i|}
\sum_{\mathbf{p}_k\in\mathcal{O}_i}
r_k^2(\bm{\Theta}^{\mathrm{sq}})\|\tilde{\mathbf{p}}_k\|.
\label{equ_sq_error}
\end{equation}

The distance-related weighting is consistent with the implementation and reduces the influence of near-center numerical instability.

The initial pose is estimated from the PCA frame of the object cluster. Robust local bounds are computed in this frame using percentile statistics, and the resulting box provides the initial center and axes. The parameters are optimized with Levenberg--Marquardt by solving
\begin{equation}
(\mathbf{J}^{T}\mathbf{J}+\lambda\mathbf{I})\Delta\bm{\uptheta}
=-\mathbf{J}^{T}\mathbf{r},
\label{equ_lm_update}
\end{equation}
where $\mathbf{J}$ is the numerical Jacobian. The candidate with the lowest residual is accepted when its fitting error and validity checks satisfy their thresholds. Otherwise, the semantic object is represented by its original GMM components.

\begin{figure}[t]
\centering
\includegraphics[width=0.98\columnwidth]{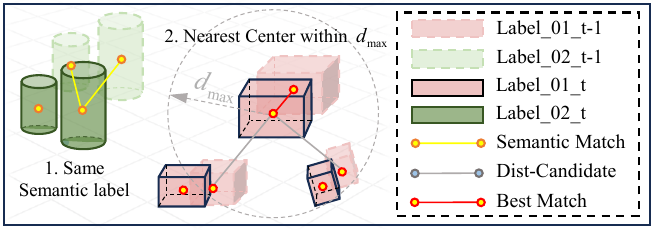}
\caption{Inter-frame superquadric association. A previous superquadric is considered only when its semantic label matches that of the current measurement. Among label-consistent candidates whose center distance is no greater than $d_{\max}$, the nearest center is selected as the correspondence used for temporal pose alignment and smoothing.}
\label{fig:sq_interframe_association}
\end{figure}

\textbf{Inter-frame association and pose stabilization.}
Temporal processing is applied only to accepted superquadric measurements. Let $Q_{m,t}$ be a current superquadric with measured center $\hat{\mathbf{c}}_{m,t}$ and semantic label $\hat{\ell}_{m,t}$, and let $\mathcal{T}^{sq}_{t-1}$ denote the previous superquadric tracks. Consistent with Fig.~\ref{fig:sq_interframe_association}, candidate tracks must have the same semantic label and lie within the maximum center distance $d_{\max}$:

\begin{equation}
\begin{aligned}
\mathcal{T}^{sq}_{m,t}=\big\{h\in\mathcal{T}^{sq}_{t-1}\;\big|\;&
\ell_h=\hat{\ell}_{m,t},\\
&\|\hat{\mathbf{c}}_{m,t}-\mathbf{c}_{h,t-1}\|_2\leq d_{\max}\big\}.
\end{aligned}
\label{equ_sq_association_candidates}
\end{equation}
The nearest valid track is selected as
\begin{equation}
h^{\ast}=\arg\min_{h\in\mathcal{T}^{sq}_{m,t}}
\|\hat{\mathbf{c}}_{m,t}-\mathbf{c}_{h,t-1}\|_2.
\label{equ_sq_association}
\end{equation}
When $\mathcal{T}^{sq}_{m,t}$ is empty, a new superquadric track is initialized. For a matched track, the center is updated by
\begin{equation}
\mathbf{c}_{h,t}=\alpha_c\hat{\mathbf{c}}_{m,t}+
(1-\alpha_c)\mathbf{c}_{h,t-1}.
\label{equ_sq_center_smoothing}
\end{equation}

Direct smoothing of the measured rotation remains unreliable because a superquadric has multiple equivalent local-coordinate representations. Flipping two local axes by $180^{\circ}$, or swapping the local $x$ and $y$ axes together with their axis lengths, can describe the same box-like object while producing a substantially different rotation matrix. Quaternion sign correction resolves only the equivalence between $\mathbf{q}$ and $-\mathbf{q}$. It does not resolve local-axis flips or permutations.

\begin{figure*}[!t]
\centering
\includegraphics[width=0.98\textwidth]{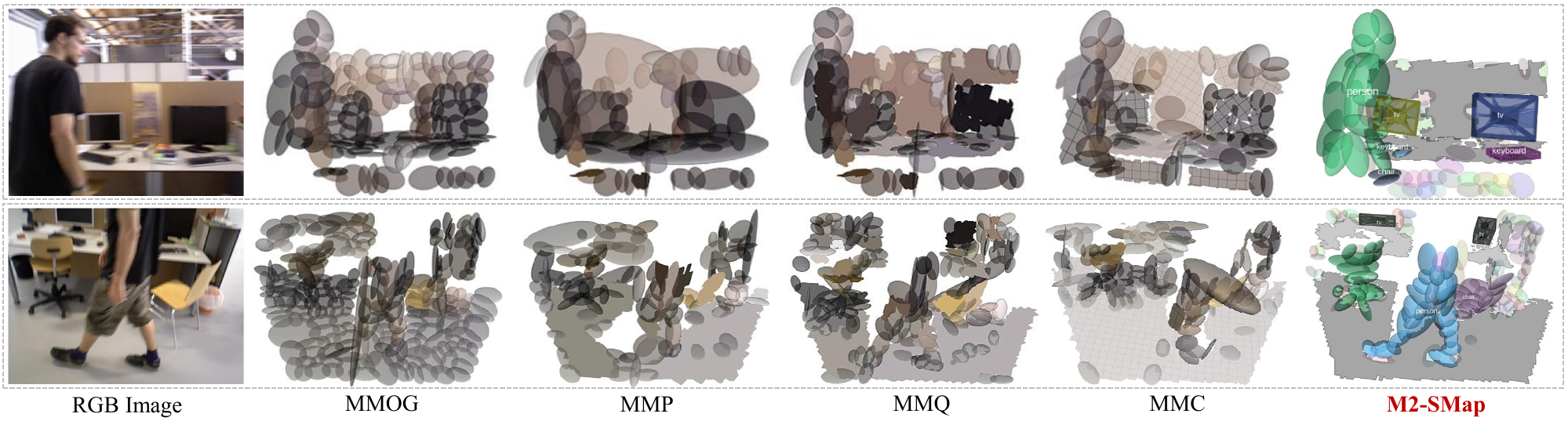}
\caption{Qualitative comparison of compact map representations on two representative RGB-D frames. Each row shows, from left to right, the input RGB image and the reconstruction results of M2-SMap, MMC, MMQ, MMP, and MMOG.}
\label{fig_existing_comparison_visual}
\end{figure*}

Given the previous matched state $(\mathbf{R}_{t-1},\mathbf{s}_{t-1})$ and the current measurement $(\hat{\mathbf{R}}_t,\hat{\mathbf{s}}_t)$, M2-SMap constructs the equivalence set
\begin{equation}
\begin{aligned}
\mathcal{E}=\{ &(\hat{\mathbf{R}}_t\mathbf{P}\mathbf{D},\mathbf{P}^{T}\hat{\mathbf{s}}_t)\; |\;
\mathbf{P}\in\{\mathbf{I},\mathbf{P}_{xy}\},\\
&\mathbf{D}=\mathrm{diag}(d_x,d_y,d_z),\;d_x,d_y,d_z\in\{-1,1\},\\
&\det(\mathbf{P}\mathbf{D})=1\},
\end{aligned}
\label{equ_sq_equiv_set}
\end{equation}
where $\mathbf{P}_{xy}$ swaps the local $x$ and $y$ axes, and $\mathbf{D}$ enumerates valid local-axis sign changes while retaining a proper rotation. The aligned measurement is selected by
\begin{equation}
(\mathbf{R}_t^{a},\mathbf{s}_t^{a})=
\arg\max_{(\mathbf{R},\mathbf{s})\in\mathcal{E}}
\mathrm{tr}(\mathbf{R}_{t-1}^{T}\mathbf{R})
-\beta\|\mathbf{s}_{t-1}-\mathbf{s}\|_2^2.
\label{equ_sq_alignment}
\end{equation}

The trace term favors orientation continuity, whereas the second term penalizes an inconsistent axis permutation.

After alignment, rotation is smoothed by quaternion SLERP using the shortest-path $\mathbf{q}/-\mathbf{q}$ correction. The axes and shape exponents are smoothed by exponential moving averages:
\begin{equation}
\begin{aligned}
\mathbf{R}_{t} &= \mathrm{slerp}(\mathbf{R}_{t-1},\mathbf{R}_{t}^{a},\alpha_R),\\
\mathbf{s}_{t} &= \alpha_s\mathbf{s}_{t}^{a}+(1-\alpha_s)\mathbf{s}_{t-1},\\
\bm{\upepsilon}_{t} &= \alpha_s\widehat{\bm{\upepsilon}}_{t}+
(1-\alpha_s)\bm{\upepsilon}_{t-1}.
\end{aligned}
\label{equ_sq_smoothing}
\end{equation}
This superquadric-specific association, alignment, and smoothing suppresses numerical pose discontinuities while preserving genuine temporal changes.

\subsection{GMM Primitives for Fallback and Residual Structures}

Semantic objects rejected by superquadric fitting and components in the residual set $\mathcal{C}_{res}$ are represented by GMM primitives. Each primitive $G_i=(\bm{\upmu}_i,\bm{\Sigma}_i,\mathbf{R}_i,\mathbf{d}_i,N_i,\mathbf{c}_i,a_i,\ell_i,s_i,\rho_i)$ stores its mean vector $\bm{\upmu}_i$, covariance matrix $\bm{\Sigma}_i$, eigenvector-based rotation matrix $\mathbf{R}_i$, principal-scale axis vector $\mathbf{d}_i$, point count $N_i$, color vector $\mathbf{c}_i$, and available semantic attributes $(a_i,\ell_i,s_i,\rho_i)$. This representation preserves local flexibility for partial observations, articulated objects, clutter, and shapes that are not well explained by a single analytic primitive. For visualization and evaluation, samples are reconstructed from each Gaussian using the covariance eigenstructure and a Mahalanobis-distance range.

\begin{figure}[!b]
\centering
\includegraphics[width=0.98\columnwidth]{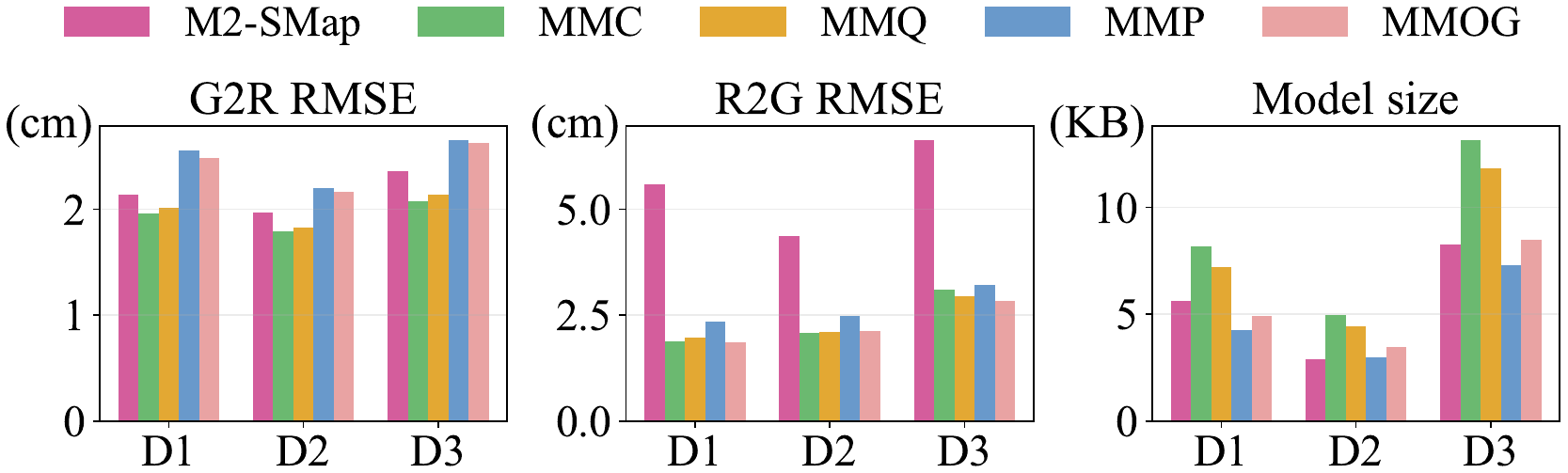}
\caption{Quantitative comparison on D1--D3. The panels report generated-to-real RMSE, real-to-generated RMSE, and mean per-frame compact model size for M2-SMap and four geometry-driven baselines.}
\label{fig:comparison}
\end{figure}

\section{Experiments}
\label{sec:experiments}

We use three TUM RGB-D sequences~\cite{sturm2012benchmark}: \textit{fr3/walking\_rpy} (D1), \textit{fr1/xyz} (D2), and \textit{fr1/teddy} (D3). They contain flat planes, planar semantic objects, regular instances, and irregular structures. YOLOv8s-seg provides instance masks, although the framework accepts any mask-level front-end. All methods use the same timestamp-aligned frames and outlier-removal protocol.

\textbf{Parameter settings.} Depth is limited to $0.1$--$3.0$~m and randomly sampled at $15\%$. The supervoxel voxel and seed resolutions are $0.03$ and $0.65$~m. Mask association requires at least $20$ valid projected points, a support ratio above $0.6$, and a YOLOv8s-seg confidence of $0.5$. Gaussian fusion uses $d_{\mathrm{connect}}=0.15$~m, $\theta_{\min}=15^{\circ}$, and $l_{\min}=0.10$~m. Flat-plane selection uses $N_{\min}=200$, $\tau_t=0.04$~m, $\tau_{\lambda}=0.02$, and $\tau_s=0.1125$~m. Superquadric fitting requires at least $80$ points, accepts fitting errors below $0.20$, and runs at most $12$ LM iterations on $100$ sampled points. Inter-frame association uses $d_{\max}=0.35$~m, $\alpha_c=0.6$, and $\alpha_s=\alpha_R=0.35$. All comparison and ablation variants share these settings unless the evaluated module is explicitly disabled.

Let $\mathcal{P}$ and $\hat{\mathcal{P}}$ denote the observed and reconstructed point clouds. The directional nearest-neighbor errors $E_{\mathrm{G2R}}=d(\hat{\mathcal{P}}\!\rightarrow\!\mathcal{P})$ and $E_{\mathrm{R2G}}=d(\mathcal{P}\!\rightarrow\!\hat{\mathcal{P}})$ measure surface accuracy and observation coverage, respectively. We also report the mean serialized primitive payload $\bar{S}_{\mathrm{map}}$, processing rate, and primitive composition.

\begin{figure*}[!t]
\centering
\includegraphics[width=0.98\textwidth]{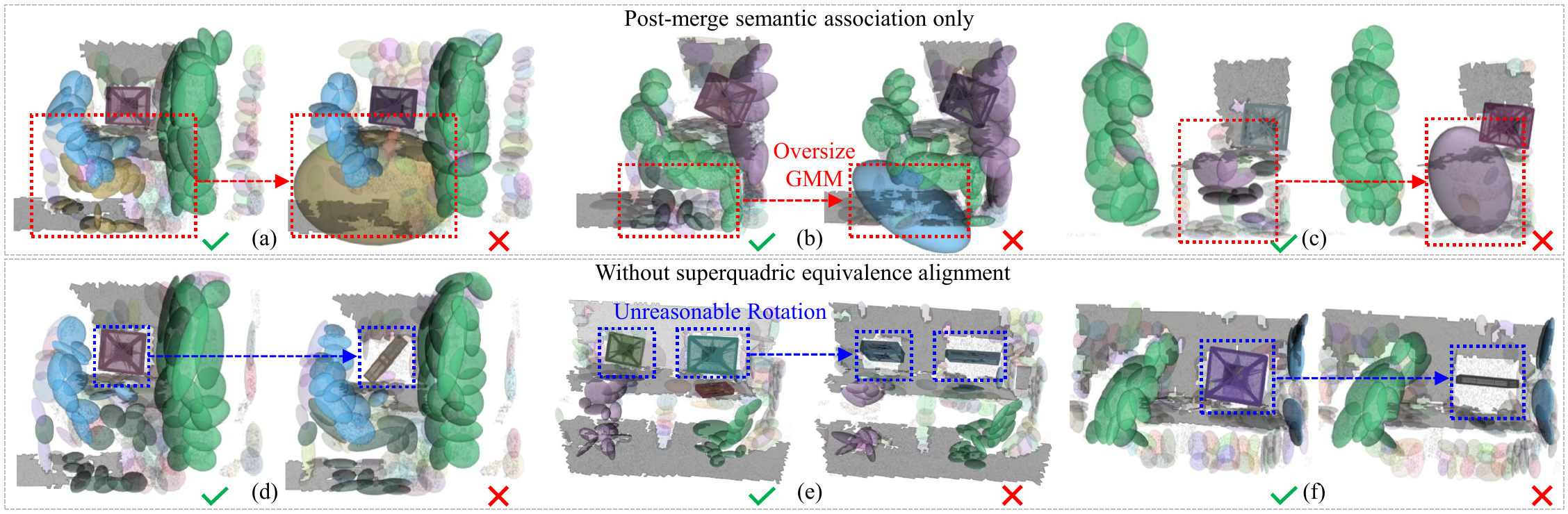}
\caption{Qualitative ablations on D1--D3. Red boxes show adhesion after geometry-only fusion, while blue boxes show spurious inter-frame SQ rotations without equivalence alignment.}
\label{fig:ablation}
\end{figure*}

\subsection{Comparison With Existing Map Representations}
\label{subsec:exp_existing_comparison}

M2-SMap is compared with four geometry-driven baselines implemented in the same framework. MMC is the complete plane--B-spline--GMM method of Gao and Dong~\cite{gao2025multimodel}. MMQ, MMP, and MMOG are controlled variants derived from this implementation by restricting the available primitive families to plane--quadric--GMM, plane--GMM, and GMM, respectively.

Fig.~\ref{fig_existing_comparison_visual} shows that M2-SMap represents object instances with coherent primitives and identifiable semantic labels, whereas the geometry-driven baselines produce less distinct object boundaries and identities.

Fig.~\ref{fig:comparison} reports the quantitative comparison. M2-SMap obtains G2R RMSEs of $0.0214$, $0.0197$, and $0.0236$~m on D1--D3. These intermediate values result from using object-level abstractions rather than the locally exhaustive fitting of MMC/MMQ or the coarser approximation of MMP/MMOG. Its R2G RMSEs are $0.0560$, $0.0437$, and $0.0663$~m, higher than the baselines because bounded planes and superquadrics intentionally avoid covering sensor noise, isolated outliers, and fine irregularities with additional local primitives.

The mean payloads are $5.613$, $2.881$, and $8.264$~KB. M2-SMap is smallest on D2 and remains close to the minimum on D1 and D3 because planes and superquadrics replace many local components, while semantic attributes and GMM fallback add limited storage. The corresponding processing rates are $97.67$, $79.90$, and $29.37$~Hz. Although semantic annotation and superquadric fitting cost more than geometry-only processing, all sequences remain real-time.

\begin{table}[!b]
\centering
\caption{Average primitive composition per aligned frame.}
\label{tab_primitive_comparison}
\setlength{\tabcolsep}{2.5pt}
\renewcommand{\arraystretch}{1.15}
\begin{tabular*}{\columnwidth}{@{\extracolsep{\fill}}ccccccc@{}}
\toprule
Method & Dataset & Plane & SQ & GMM & B/Q & Total \\
\midrule
\multirow[c]{3}{*}{M2-SMap}
 & D1 & 1.348 & 0.804 & 49.478 & 0.000 & \textbf{51.630} \\
 & D2 & 1.193 & 1.207 & 24.281 & 0.000 & \textbf{26.682} \\
 & D3 & 2.353 & 1.176 & 72.529 & 0.000 & \textbf{76.059} \\
\midrule
\multirow[c]{3}{*}{MMC}
 & D1 & 3.870 & 0.000 & 50.217 & 4.870 & 58.957 \\
 & D2 & 2.949 & 0.000 & 32.151 & 1.810 & 36.909 \\
 & D3 & 5.471 & 0.000 & 78.824 & 6.765 & 91.059 \\
\midrule
\multirow[c]{3}{*}{MMQ}
 & D1 & 3.783 & 0.000 & 50.478 & 4.848 & 59.109 \\
 & D2 & 3.020 & 0.000 & 32.054 & 1.810 & 36.884 \\
 & D3 & 5.765 & 0.000 & 80.941 & 6.647 & 93.353 \\
\midrule
\multirow[c]{3}{*}{MMP}
 & D1 & 4.065 & 0.000 & 54.544 & 0.000 & 58.609 \\
 & D2 & 3.026 & 0.000 & 33.892 & 0.000 & 36.918 \\
 & D3 & 6.000 & 0.000 & 86.118 & 0.000 & 92.118 \\
\midrule
\multirow[c]{3}{*}{MMOG}
 & D1 & 0.000 & 0.000 & 78.674 & 0.000 & 78.674 \\
 & D2 & 0.000 & 0.000 & 55.276 & 0.000 & 55.276 \\
 & D3 & 0.000 & 0.000 & 135.765 & 0.000 & 135.765 \\
\bottomrule
\end{tabular*}
\end{table}

Table~\ref{tab_primitive_comparison} reports primitive composition, where B/Q denotes B-splines in MMC and quadrics in MMQ. M2-SMap achieves the lowest total count on all sequences, reducing the best-baseline count by $11.9\%$, $27.7\%$, and $16.5\%$ on D1--D3. The reduction comes from representing large planar supports and regular objects with single bounded planes and superquadrics, while retaining GMMs only for irregular or superquadric-rejected structures.

\subsection{Ablation Study}
\label{subsec:exp_ablation}

The ablations compare the full method with post-merge semantic association only (PMSAO) and the variant without superquadric equivalence alignment (WSQA), using identical inputs and parameters. $A_s$, $N_{adh}$, and $J_{sq}$ denote semantic fusion consistency, adhesion count, and mean inter-frame SQ rotation jump, respectively. The semantic plane-exclusion rule remains enabled in all variants. Fig.~\ref{fig:ablation} and Table~\ref{tab_ablation} report the results.

\begin{table}[t]
\centering
\caption{Ablation results on three RGB-D sequences.}
\label{tab_ablation}
\setlength{\tabcolsep}{2.7pt}
\renewcommand{\arraystretch}{1.15}
\begin{tabular*}{\columnwidth}{@{\extracolsep{\fill}}ccccccc@{}}
\toprule
Method & Dataset & $R_c$ & $T$(ms) & $A_s$ & $N_{adh}$ & $J_{sq}$($^\circ$) \\
\midrule
\multirow[c]{3}{*}{Full}
 & D1 & 9.174 & 41.281 & 1.000 & 0.000 & 10.404 \\
 & D2 & 11.578 & 15.512 & 1.000 & 0.000 & 8.163 \\
 & D3 & 10.368 & 54.985 & 1.000 & 0.000 & 14.725 \\
\midrule
\multirow[c]{3}{*}{PMSAO}
 & D1 & 9.375 & 41.481 & 0.810 & 4.281 & 10.546 \\
 & D2 & 12.518 & 15.528 & 0.516 & 1.851 & 8.258 \\
 & D3 & 10.455 & 54.910 & 0.777 & 2.291 & 14.520 \\
\midrule
\multirow[c]{3}{*}{WSQA}
 & D1 & 9.307 & 40.112 & 1.000 & 0.000 & 61.213 \\
 & D2 & 11.571 & 15.474 & 1.000 & 0.000 & 43.574 \\
 & D3 & 10.386 & 54.385 & 1.000 & 0.000 & 50.058 \\
\bottomrule
\end{tabular*}
\end{table}

\textbf{Post-merge semantic association only.}
Without pre-merge semantics, $A_s$ decreases from $1.000$ to $0.810$, $0.516$, and $0.777$, while $N_{adh}$ increases from zero to $4.281$, $1.851$, and $2.291$ on D1--D3. Processing changes by less than $0.21$~ms. The slightly higher $R_c$ is not a true improvement because incorrect fusion produces fewer but oversized components, and post-merge labels cannot recover the lost object--plane or inter-object boundaries.

\textbf{Without superquadric alignment.}
Disabling equivalence alignment increases $J_{sq}$ from $10.404^{\circ}$ to $61.213^{\circ}$ on D1, $8.163^{\circ}$ to $43.574^{\circ}$ on D2, and $14.725^{\circ}$ to $50.058^{\circ}$ on D3. The full method therefore reduces the jumps by $83.0\%$, $81.3\%$, and $70.6\%$. The nearly unchanged $R_c$, $A_s$, $N_{adh}$, and processing time show that alignment specifically suppresses equivalent-pose discontinuities rather than affecting compactness or semantic grouping.

\section{CONCLUSIONS}

Existing compact multi-model mapping methods are often dominated by geometric criteria, which can lead to inefficient primitive allocation and inter-object adhesion. To address these limitations, we propose M2-SMap, a memory-efficient semantic mapping framework that integrates instance-level semantic constraints with hierarchical multi-model representation. By combining bounded planes, object-level superquadrics, and GMM primitives, M2-SMap improves representation compactness while preserving semantic consistency. Experiments on three RGB-D sequences show that M2-SMap reduces the average primitive count by 18.7\% over the best baseline, eliminates the measured inter-object adhesion cases from 2.808 to 0, and maintains real-time processing at 29.37-97.67~Hz. Future work will investigate larger-scale and dynamic environments and tighter integration with robotic navigation and manipulation.

\bibliographystyle{IEEEtran}
\bibliography{m2smap_references}

\end{document}